# Automated Feedback Loops to Protect Text Simplification with Generative AI from Information Loss

Abhay Kumara Sri Krishna Nandiraju[1], Gondy Leroy[1], David Kauchak[2], and Arif Ahmed[1]

[1]University of Arizona, Tucson, USA
[2]Pomona College, Claremont, USA
```
{abhaynandiraju, gondyleroy, arifahmed}@arizona.edu
             david.kauchak@pomona.edu
```

**Abstract.** Understanding health information is essential in achieving and maintaining a healthy life. We focus on simplifying health information for better understanding. With the availability of generative AI, the simplification process has become efficient and of reasonable quality, however, the algorithms remove information that may be crucial for comprehension. In this study, we compare generative AI to detect missing information in simplified text, evaluate its importance, and fix the text with the missing information. We collected 50 health information texts and simplified them using gpt-4-0613. We compare five approaches to identify missing elements and regenerate the text by inserting the missing elements. These five approaches involve adding missing entities and missing words in various ways: 1) adding all the missing entities, 2) adding all missing words, 3) adding the top-3 entities ranked by gpt-4-0613, and 4, 5) serving as controls for comparison, adding randomly chosen entities. We use cosine similarity and ROUGE scores to evaluate the semantic similarity and content overlap between the original, simplified, and reconstructed simplified text. We do this for both summaries and full text. Overall, we find that adding missing entities improves the text. Adding all the missing entities resulted in better text regeneration, which was better than adding the top-ranked entities or words, or random words. Current tools can identify these entities, but are not valuable in ranking them.

## 1    Introduction

It is increasingly important to be able to retrieve and evaluate health information. Much information, as well as misinformation, is available on the Internet, and a variety of AI tools generate information largely using that Internet knowledge as a baseline. They are largely unchecked.

Over the last few decades, a variety of tools have been developed and promoted to leverage information technology [1-6]. Early work included the development of readability formulas, and although they have not been shown to increase information understanding, they have been integrated into text editors and are extremely popular [7, 8, 9]. More recently, data-driven approaches have been proposed that focus on specific text or audio characteristics that are related to better comprehension and retention of information [10-19]. While these approaches have been shown to improve understanding, they are not automated and require a human-in-the-loop. In the last few years, AI tools have become available that can simplify, summarize, or generate health-related information [3-6]. These tools are powerful but are not guaranteed to provide correct or complete information [20, 21].

Building on our prior work with text simplification [13, 15, 16-19], we present an approach that leverages generative AI (ChatGPT) to simplify text and leverages an automated feedback loop to ensure no critical information is omitted.

## 2     Related Work

### 2.1    Information Distribution in Healthcare

Artificial Intelligence (AI) has evolved dramatically over recent decades, with generative AI emerging as one of the most transformative technological developments of the early 21st century. Generative AI refers to artificial intelligence systems capable of creating new content, including text, images, audio, and more [1]. It has significantly advanced text generation across multiple applications, including content creation, dialogue systems, creative writing assistance, text summarization, and text simplification [3-6].

Large Language Models (LLMs), such as GPT-4, can generate coherent essays and technical content and are also used to summarize and simplify text content [2]. Text summarization or simplification using Large Language Models (LLMs) has emerged as a powerful tool. However, it faces significant challenges related to content deletion and information retention. Tariq et al [21] identified deletion errors in ChatGPT-generated summaries, referring to the omission of critical information when summarizing complex texts. Their analysis showed that approximately 21 per cent of ChatGPT-generated simplification of a given medical text omitted key facts present in the original text. Tang et al [19] examined ChatGPT's performance in summarizing medical literature and found significantly higher deletion rates (28 to 35 per cent) compared to general texts (15 to 20 per cent). They emphasized that this issue poses serious risks in clinical applications, where missing details could lead to misinformation or adverse outcomes.

We propose a novel approach that combines an automated feedback loop to effectively tackle the deletion problem for healthcare text data using ChatGPT. Our method focuses on identifying the missing information in simplified texts and, using a feedback loop, reinserting the missing details into the simplified version, ensuring a more accurate and refined output. This process not only mitigates the loss of important information but also generates a significantly improved simplified text that retains the meaning and context of the original. Through this innovative approach, we aim to enhance the overall quality of simplified text generation.

## 2.2  Generative AI for Information Simplification

Generative AI is being evaluated to address many natural language understanding tasks such as textual entailment, question answering, semantic similarity, and document classification. The GPT-1 model outperformed other discriminative models, such as task-specific architectures, on various NLP tasks including Natural Language Inference, Question Answering, and Text Classification [22]. These generative pre-training methods gave rise to language models capable of understanding the semantics and generating new words based on the language structures in each input sequence [22]. Moreover, pre-training enables the model to be fine-tuned later for specific tasks to achieve state-of-the-art performance while still maintaining the knowledge from pre-training. Later, with the development of GPT-3 [23], it was identified that scaling up autoregressive language models enabled the models to perform competitively across a wide variety of tasks with minimal supervision. In addition to this, such models were able to adapt to new settings via few-shot prompting techniques, but few-shot learning strategies still struggled on certain domain-specific tasks such as text simplification.

With proper supervision [24], the usage of generative AI methods in healthcare can save time in preparing crucial biomedical reports and documents. For example, discharge summaries [24] are documents containing detailed information about the patient's condition, ongoing diagnosis, and suggested treatments to improve their health situation. These discharge summaries are traditionally time-consuming, burdening doctors and risking delays in patient care. The text-generation ability of Large Language Models(LLMs) can be used to generate formal discharge summaries if they are provided with specific information about patients, reducing the burden on doctors. However, it should be noted that the discharge summaries generated by LLMs like ChatGPT are prone to errors [24] and hallucinations since they are not fine-tuned to understand complex biomedical terms. To solve this issue, GPT-3.5-turbo-16k was first used by Stanceski et al. [24] to simplify the discharge summaries based on a prompt that balances language simplification along with correctness of the medications [24]. Zaretsky et al.[25], found that LLMs can also translate complex discharge summaries into patient-friendly language. However, important details went missing along with the introduction of hallucinations in the generated discharge summaries [24]. Similarly, it was found that patient clinic letters generated by LLMs were more readable, but they omitted certain important details [26].

Large Language Models are good at simplifying complex material, but often face a trade-off between simplification and information retention. While LLMs produce more human-readable text, they often omit crucial details and context from the original text [27]. This poses a problem with complex biomedical information related to medicines, diagnosis, and treatments. In addition to this, LLMs are also prone to produce incorrect information that is factually wrong or ambiguous to evaluate based on the original source information [28]. In our work, we automatically identify missing information in a simplified text and add it back to the simplified text to generate an improved simplified version.

## 3      Text Simplification Without Information Omission

Our goal is to simplify complex biomedical texts without information loss. To do this we evaluate an automatic feedback loop comprising three steps: text simplification, identification of missing elements, and then regeneration of the simplified text, with specific guidance regarding the missing elements. Identifying missing elements involves finding missing entities and missing words. Regeneration of simplified text is performed by inserting the missing elements back and generating an improved simplified version. We compare five different approaches to insert the missing information. A detailed description of identifying missing elements and simplified text regeneration is included below. We use automated metrics, cosine similarity and ROUGE-1 scores, to evaluate the semantic similarity and content overlap between the original text and the generated simplified text across all five approaches. We assume for this study that a higher similarity to the original text reflects more complete information.

### 3.1      Missing Information Identification

We identify the missing information between the original text and the simplified text using two approaches.

The first approach identifies words occurring frequently in the original text but are less frequent in the simplified text. Text is tokenized, stop words are removed, and all the words are stemmed using NLTK[*]. We identify missing words as tokens present in the original text two or more times, but less than twice in the simplified text. Let $T_{original}$ and $T_{simplified}$ represent the original and simplified texts respectively, and after preprocessing, let $W_{original}$ and $W_{simplified}$ represent the sets of words in the original and simplified texts. Now the set of missing words $W_{missing}$ is defined as:

$$W_{missing} = \{w \in W_{original} | f(w, T_{original}) \geq 2 \text{ and } f(w, T_{simplified}) < 2\}$$

where, $f(w, T)$ denotes the frequency of word w in text T.

The second approach focuses on identifying missing named entities. We extracted the biomedical named entities using scispacy[†] from the original and simplified text, and the set-theoretic subtraction operation is performed to identify missing named entities. Let $SE_{original}$ represent the set of entities from the original text and $SE_{simplified}$ represent the set of entities from the simplified text. $SE_{missing}$ is defined as the difference between the two sets:

$$SE_{missing} = SE_{original} \setminus SE_{simplified}$$

where, \ denotes the set difference operation, resulting in named entities that belong to the original text($SE_{original}$) but not to the simplified text($SE_{simplified}$).

---

[*] Link to the nltk library used for text pre-processing: https://www.nltk.org/
[†] Link to the scispacy library used for name entity recognition: https://allenai.github.io/scispacy/

### 3.2    Text Regeneration

Once the missing named entities and missing words were identified, they are inserted back into the simplified text using gpt-4-0613[‡] model with the prompt in Fig. 1:

```
prompt = f"""Task: Create an improved simplified version of a text that
incorporates missing important entities while maintaining simplicity.

Context:
- Original Text: {original_text}
- Current Simplified Text: {simplified_text}
- Important Entities Missing: {missing_entities}

Instructions:
1. Start with the current simplified text as your base
2. Incorporate the missing entities naturally and contextually
3. Maintain the simplified language level of the current simplified text
4. Ensure the text flows smoothly and logically
5. Keep the meaning as close as possible to the original text
6. Do not add information that isn't in the original text
7. Use the missing entities in their proper context from the original text

Please provide only the new simplified text without any explanations or
additional notes.
"""
```

**Fig. 1.**  The figure shows a structured prompt guiding the augmentation of simplified text with missing entities or words while retaining original context and linguistic simplicity

We evaluate five different approaches (A1 to A5) to insert the missing information back into the text to generate an improved simplified text that has better semantic similarity and content overlap with the original text. In each of the five approaches, the missing information was added to the simplified text to generate an augmented simplified version. The approaches used are:

1) A1: All the missing entities ($SE_{\text{missing}}$) are added to the simplified text to generate an improved simplified version by providing the corresponding details to the prompt mentioned above.
2) A2: All the missing words ($W_{missing}$) are added to the simplified text to generate an improved simplified version.
3) A3: We used gpt-4-0613 model to rank the missing entities for importance in the full text. The top-3 ranked entities are added to the simplified text to

---

[‡] Link to the gpt-4-0613 model used for text simplification:
https://platform.openai.com/docs/models/gpt-4

generate an improved version. The prompt used for ranking the entities is shown in Fig. 2:

```
prompt = f"""Task: Analyze and rank missing entities based on their importance to
the text.

Context:
- Original Text: {original}
- Current Simplified Text: {simplified}
- Missing Entities: {', '.join(entities)}

Instructions:
1. Analyze the importance of each missing entity to the overall meaning of the text
2. Rank the entities from most important to least important
3. Select the top 3 most crucial entities that would add the most value to the
simplified text

Please provide:
1. A ranked list of all missing entities from most to least important
2. The top 3 most important entities

Format your response as a JSON object with these keys:
- ranked_entities: [list of all entities ranked by importance]
- top_3_entities: [list of the 3 most important entities]
"""
```

**Fig. 2.** The figure shows a prompt guiding the evaluation and selection of the most semantically valuable entities to enhance the simplified text.

Approaches 4 and 5 are two control conditions in which we insert random information. These are necessary to verify that adding any information does not result in increased similarity.

1) A4: three random entities are chosen randomly from the set of missing entities ($SE_{missing}$) and are added to the simplified text to generate an improved version.
2) A5: k-random entities are chosen from the set of missing entities ($SE_{missing}$) and added to the old simplified text to generate an improved version, where k is the cardinality of the set $W_{missing}$ i.e., the number of missing words that are present in the original text at least twice but less than twice in the simplified text.

### 3.3 Regenerated Text Evaluation

We evaluate the final text using cosine similarity and ROUGE-1 [29] score to assess the semantic similarity and content overlap with the original text. We apply both metrics on full text and summaries of the text. We generate summaries to evaluate whether the final text retains the core meaning of the original text, because the summary is a condensed representation containing the core aspects and details of the original text. This two-step evaluation provides a comprehensive analysis of the final text, allowing us to capture both semantic similarity and content overlap or lexical similarity in both original and condensed representations.

The summaries of the original text, simplified text and augmented simplified text are generated using a BART [30] model. Bidirectional and Auto-Regressive Transformer (BART) is a sequence-to-sequence model pre-trained as a denoising autoencoder [30]. Since it is trained to reconstruct the original text from a noisy input text [30], it is apt for text generation tasks such as summarization, machine translation, and question answering. Hence, BART was chosen for the summarization task.

To evaluate the generated texts, cosine similarity and ROUGE-1[29] scores were computed to compare the original text and the new simplified text, and between the summaries of the original text and the new simplified text. This process is repeated for all five approaches.

## 4  Evaluation

### 4.1  Data Set

We collected rheumatology-based texts from the British Medical Journal (BMJ) (N=663), which were shown in our prior work to be very complex for lay people to understand [18]. We randomly selected 50 texts, identified the missing entities and missing words, and inserted them back into the simplified text to generate an augmented simplified version using the five approaches discussed earlier. The respective summaries were developed using the BART [30] model, and the metrics were computed for the full texts and the summaries of full texts. The metric computations are discussed in detail in the section below.

### 4.2  Metrics

Cosine similarity is an effective way to measure the similarity between two texts. It measures the cosine of the angle between the vector representations of the two texts(vector embeddings) in a high-dimensional space. Initially, each text is represented as a vector embedding in a 384-dimensional dense vector space using the embedding model all-MiniLM-L6-v2 from the sentence-transformers library. After representing the texts as embedding vectors, the cosine of the angle between these vectors is computed to find the cosine similarity between the two texts. Let $T_1$ and $T_2$ represent two texts, and let their 384-dimensional vector representations be denoted by A and B, respectively. The cosine similarity is computed as:

$$cos(T_1, T_2) = \frac{\boldsymbol{A} \cdot \boldsymbol{B}}{|\boldsymbol{A}||\boldsymbol{B}|}$$

where,

$$\boldsymbol{A} \cdot \boldsymbol{B} = \sum_{i=1}^{n=384} A_i B_i,$$

$$|\boldsymbol{A}| = \sqrt{\sum_{i=1}^{384} A_i^2},$$

$$|\boldsymbol{B}| = \sqrt{\sum_{i=1}^{384} B_i^2},$$

Using the above approach $cos(T_{\text{original}}, T_{\text{simplified}})$, $cos(T_{\text{original}}, T_{\text{augmented simplified}})$, $cos(S_{\text{original}}, S_{\text{simplified}})$, and $cos(S_{\text{original}}, S_{\text{augmented simplified}})$ were computed, where S denotes the summary of a text T. The value is 1 if two texts are identical and is 0 if they are completely dissimilar. The cosine similarity measure indirectly reflects the semantic relationship or similarity between texts because the text embeddings are designed to capture contextual and semantic information in a given text.

Another standard metric to assess text similarity is the ROUGE [29] score, especially in tasks like simplification and summarization. It measures the overlap between the words or sequence of words between a candidate text (simplified text) and a reference text (original text) [30]. In our case, we use the ROUGE-1 score since it focuses on the overlap of unigrams [30] between two texts, resulting in a metric that helps us in assessing how much of the content in terms of individual words in the original text is retained in the simplified text [30]. Apart from the content retention, it also indicates the relevance of simplified text to the original text, providing a balance between retention and relevance. Let To and Ts represent the original (reference) and simplified (candidate) texts, respectively, and ROUGE-1 score or ROUGE-1 F1-score is defined as the harmonic mean of ROUGE-1 recall and ROUGE-1 precision. Let R1-r, R1-p and R1-f1 represent the ROUGE-1 recall, ROUGE-1 precision and ROUGE-1 score/ROUGE-1 F1-score respectively. They are computed as:

$$R1 - r_{T_0, T_S} = \frac{|the\ U_{\text{overlap}}|}{|U_{T_0}|}$$

$$R1 - p_{T_0, T_S} = \frac{|U_{\text{overlap}}|}{|U_{T_S}|}$$

$$R1 - f1_{T_0, T_S} = \frac{2 \cdot R1 - r_{T_0, T_S} \cdot R1 - p_{T_0, T_S}}{R1 - r_{T_0, T_S} + R1 - p_{T_0, T_S}}$$

where,

$|U_{T_o}|$: cardinality of the set of unigrams in the original text($T_o$)

$|U_{T_s}|$: cardinality of the set of unigrams in the simplified text($T_s$)

$|U_{overlap} = U_{T_o} \cap U_{T_s}|$: cardinality of the set of overlapping unigrams

Using the above discussed approach $R1 - f1_{\{T_{original}, T_{old\ simplified}\}}$, $R1 - f1_{\{T_{original}, T_{new\ simplified}\}}$, $R1 - f1_{\{S_{original}, S_{old\ simplified}\}}$, $R1 - f1_{\{S_{original}, S_{new\ simplified}\}}$ are computed where S denoted the summary of a text T.

Comparing the results at the document level (original vs simplified) and at the summary level (summaries of original vs simplified) provides a comprehensive two-step analysis of the texts at original representation and the condensed representation. This allows us to draw conclusions more effectively rather than looking at the results from original representations themselves.

## 5    Results

Using the original text and the simplified text, we calculated the average cosine similarity and ROUGE-1 scores for both the full text and summaries. The results are shown in Table 1. The values in Table 1 for the metrics between the original and simplified text show the initial alignment and content overlap. We see that the values are higher for full text compared to summaries.

When comparing the augmentation strategies (A1 to A5), we see that A1 achieves the highest mean cosine similarity of 0.9162, which is higher than all methods, and A2 achieves the second highest mean cosine similarity of 0.8990, which is slightly better than A5 with the mean cosine similarity of 0.8967. A3 and A4 are the worst, with a mean cosine similarity of 0.8758 and 0.8632, respectively. In terms of ROUGE-1 scores, A1 still obtains the best mean ROUGE-1 score of 0.6555, A2 obtains the mean ROUGE-1 score of 0.6243, and A5 reaches the mean ROUGE-1 score of 0.6051. A3 and A4 are the worst, with a mean ROUGE-1 score of 0.5364 and 0.5454, respectively.

This shows that A1 is the best at preserving the meaning and the content for the entire document when using the full text for comparison.

For the summaries, the best average cosine similarity score is obtained by A2 with 0.8058, followed by A5 with 0.7848 and A4 with 0.7742; thus, A3 and A1 are the worst with 0.7632 and 0.7609. As for the ROUGE-1 scores, A2 still has the best score with 0.4213, because it is higher than other methods, followed by A5 with 0.4066 and A1 with 0.3908. The worst scores are obtained by A3 and A4 with 0.3844 and 0.3758, therefore A2 is the top method for summaries, and thus this suggests that A2 is the best in capturing the fine-grained details and preserving the alignment for summaries.

This shows that A2 is the best at preserving the meaning and the content for the summarized representations when using the summaries for comparison.

When comparing the simplified and augmented simplified text, we see that similarity with the original text increases for all approaches, with A1 showing the largest improvement at the document level and A2 showing the largest improvement at the summary level. This demonstrates that adding all missing entities (A1) or all missing words (A2) is more effective than other strategies for improving semantic alignment and content overlap.

**Table 1.** Average metric values between original text and simplified text and augmented text

|    | Full Text | | Summaries | |
|----|-----------|---------|-----------|---------|
|    | Cosine Sim | ROUGE-1 | Cosine Sim | ROUGE-1 |
| No insertion: Original-simplified metrics | | | | |
|    | 0.8471 | 0.5063 | 0.7583 | 0.3528 |
| Insertion Approach: Original-Augmented simplified metrics | | | | |
| A1 | **0.9162** | **0.6555** | 0.7609 | 0.3908 |
| A2 | 0.8990 | 0.6243 | **0.8058** | **0.4213** |
| A3 | 0.8758 | 0.5364 | 0.7632 | 0.3844 |
| A4 | 0.8632 | 0.5454 | 0.7742 | 0.3758 |
| A5 | 0.8967 | 0.6051 | 0.7848 | 0.4066 |

Approaches A4 and A5 are our control conditions, where random information is inserted. As expected, we see a clear drop in performance compared to A1 and A2, particularly in terms of ROUGE-1 scores. This suggests that systematic insertion of missing entities or words (as in A1 and A2) is more effective than random insertion for preserving semantic content and alignment with the original text.

## 6   Discussion

We explored five approaches (A1 to A5) to improve the semantic alignment and content overlap between the original text and simplified versions. Each approach involved adding missing entities or words to the simplified text to regenerate and create an augmented simplified version. We evaluated performance using cosine similarity and ROUGE-1 scores at both the document level and summary level.

Overall, we found that approach A1, which adds all the missing entities, consistently performs the best at the document level, achieving the highest ROUGE-1 and cosine similarity scores. Approach A2, which adds all the missing words performs well at the summary level. A1 outperforms A2 in terms of cosine similarity and ROUGE-1 score at the document level, whereas at the summary level A2 outperforms A1. This indicates that approach A1, which focuses on adding all missing entities($SE_{missing}$) is able to capture broader semantic concepts and improves document-level alignment but A2 which focuses on adding all missing words($W_{missing}$) is able to capture fine-grained details, resulting in improved summary-level alignment. Overall, approach A1 performs better than the rest since it is able to capture semantic concepts at a deeper level compared to the rest.

Approaches A3, which adds top-3 entities ranked by gpt-4-0613, and A4, which adds three random entities underperform significantly. Adding only three entities regardless of the selection process, does not improve semantic alignment and content overlap. The inconsistencies in the patterns of mean cosine similarity values and ROUGE-1 scores across document and summary levels in approaches A3 and A4 indicate that the ranking mechanism of the gpt-4-0613 model does not provide a significant advantage over selecting the same number of entities randomly. Additionally, it also indicates that choosing three entities irrespective of the selection method does not result in any significant improvement in semantic preservation.

Similarly, approach A5, which adds k-random entities, where k is the number of missing words ($|W_{missing}|$), consistently underperforms approach A2, which adds all the missing words ($W_{missing}$) systematically. The metric values suggest that A2 is better than A5, as it consistently achieves higher cosine similarity and ROUGE-1 scores at both summary and document levels. Additionally, it suggests that adding all the missing words ensures more content overlap and semantic preservation than adding the same number of missing entities.

Approach A4 and A5, which involves adding three and k ($=|W_{missing}|$) random entities respectively underperforms as expected compared to adding all the missing entities. These strategies show that using cosine similarity and ROUGE-1 scores capture more than just added information.

## 7 Conclusion

In this study, we are identifying the missing information in the simplified text, inserting missing information back to regenerate an improved simplified version using five approaches, and evaluating and comparing the regenerated simplified version across all five approaches. Our goal is to simplify the complex biomedical texts to human-readable texts without information loss and ensure that the regenerated simplified text has the best semantic content and improves overall alignment with the original text.

Our evaluation indicated that Approach A1 (adding all missing entities) provided the best overall improvement for document-level semantic alignment, whereas approach A2 (adding all missing words) was most effective at improving summary-level alignment, likely due to its focus on finer details. Approaches that involved adding a certain number of entities (A3, A4, A5), whether selected via ranking or randomly, demonstrated significantly less improvement, highlighting that comprehensive information reinsertion is effective than partial or random additions for retaining semantic meaning. Ultimately, adding all missing entities (A1) appears to be the most effective and robust method evaluated for improving the overall semantic correctness of simplified biomedical texts.

## 8 Future Work

Our future improvements focus on exploring hybrid approaches that combine the strengths of A1 and A2, such as adding both missing entities and words to better capture broader semantic concepts as well as fine-grained details. Additionally, we will focus on alternative ranking mechanisms to pick the crucial entities and experiment with dynamic thresholds for selecting missing words or entities to optimize the balance between semantic preservation and content overlap.

## Acknowledgments

The research reported in this study was supported by the National Library of Medicine of the National Institutes of Health under Award Number R01LM011975. The content is solely the responsibility of the authors and does not necessarily represent the official views of the National Institutes of Health.